\documentclass{article}
\PassOptionsToPackage{numbers,compress}{natbib}
\usepackage[final,nonatbib]{tackling_climate_workshop_style}
\usepackage[utf8]{inputenc}
\usepackage[T1]{fontenc}
\usepackage{hyperref}
\usepackage{url}
\usepackage{booktabs}
\usepackage{amsfonts}
\usepackage{amsmath}
\usepackage{nicefrac}
\usepackage{microtype}\usepackage{tikz}
\usetikzlibrary{arrows.meta,positioning,calc}
\usepackage{xcolor}
\usepackage{graphicx}

\title{Benchmarking Segmentation Models for Scalable Mangrove Mapping with Sentinel-1/2 Imagery}

\author{
  \textbf{Damba Koné}\textsuperscript{1}
  \and \textbf{Sara Mobsite}\textsuperscript{1}
  \and \textbf{Joris Guérin}\textsuperscript{1,2,3}
  \and \textbf{Élodie Blanchard}\textsuperscript{1}
  \and \textbf{Thibault Catry}\textsuperscript{1,4}
  \\ \\
  \textsuperscript{1}UMR ESPACE-DEV, IRD, University of Montpellier, France\\
  \textsuperscript{2}Federal University of Rio Grande do Norte, Brazil\\
  \textsuperscript{3}International Joint Laboratory IDEAL, Brazil/France\\
  \textsuperscript{4}KHEOBS Laboratory, Institute of Technology of Cambodia, Phnom Penh, Cambodia\\
  \texttt{\small \{damba.kone,sara.mobsite,joris.guerin,elodie.blanchard,thibault.catry\}@ird.fr}
}

\begin{document}
\maketitle

\begin{abstract}
Mangroves are dynamic coastal ecosystems that contribute to carbon storage, coastal protection and biodiversity conservation but are vulnerable to climate change. Very-high-resolution imagery such as Pléiades enables detailed mangrove mapping but it is costly and have a low revisit rate. Large-scale monitoring requires free and regularly available data such as Sentinel-1 and Sentinel-2 but their spatial resolution poses challenges for detailed mangrove mapping. In this work, we benchmark deep learning approaches for mangrove mapping from Sentinel-1 radar and Sentinel-2 optical imagery using reference maps derived from very-high-resolution Pléiades imagery. We compare four segmentation configurations based on CROMA and BigEarthNet-pretrained ResNet50 representations with different segmentation decoders. We construct a mangrove mapping dataset covering French Guiana, Suriname, and Amapá. CROMA + UPerNet achieves the best performance, with an F1-score of \textbf{0.9218} and an IoU of \textbf{0.8549}. These results highlight the potential of pretrained multimodal Earth observation representations and open Sentinel data to support mangrove resilience through scalable monitoring.
\end{abstract}

\section{Introduction}
Mangroves are tropical and subtropical coastal ecosystems that provide important services  storing substantial amounts of carbon in their biomass and soils, supporting biodiversity, and contributing to shoreline protection \cite{alongi2002,alongi2012,tomlinson2016}. Monitoring their spatial distribution is important for understanding ecosystem dynamics, monitoring blue-carbon and supporting conservation policies. Our study focuses on French Guiana mangroves, where sediments transported from the Amazon along the coast drive alternating phases of accretion and erosion, resulting in continuous changes in mangrove extent \cite{fromard1998,fromard2004,anthony2014,augusseau2024}.

Earth observation provides a practical means of monitoring these ecosystems over vast territories and through time \cite{giri2011,bunting2018}. Very-high-resolution imagery such as Pléiades (0.5 m spatial resolution) can provide detailed information for mangrove mapping, but its cost and limited systematic availability constrain its use for repeated large-scale monitoring. In contrast, Sentinel-1 (S1) and Sentinel-2 (S2) provide freely and regularly available observations at 10~m spatial resolution. Sentinel-1 synthetic aperture radar (SAR) is largely unaffected by cloud cover and provides information related to surface and vegetation structure while Sentinel-2 provides complementary multispectral information for vegetation characterization \cite{torres2012,drusch2012}.

Previous mangrove mapping approaches based on deep learning typically rely on large annotated datasets, which are costly to produce at large spatial scales \cite{ghorbanian2021,sharifi2022,moreno2023}. Recent self-supervised Earth observation models provide an alternative by learning transferable representations from large collections of unlabeled satellite imagery. In this work, we investigate deep learning for binary mangrove segmentation from multimodal Sentinel-1 and Sentinel-2 imagery using reference maps derived from Pléiades imagery for supervision. We compare CROMA \cite{fuller2023} representations with a ResNet50 encoder pretrained on BigEarthNet \cite{clasen2025refinedbigearthnet,hackel2024configilm} and evaluate CNN, UPerNet \cite{xiao2018}, and DeepLabV3+ \cite{Chen2018} segmentation decoders. Beyond model architecture, we adopt a training strategy that incorporates temporal compositing, class-imbalance mitigation, and geographic enrichment of the training data, while using geographically separated training, validation, and test areas to limit spatial leakage.

\section{Data and methodology}
\begin{figure}[!h]
    \centering
    \includegraphics[width=0.7\linewidth]{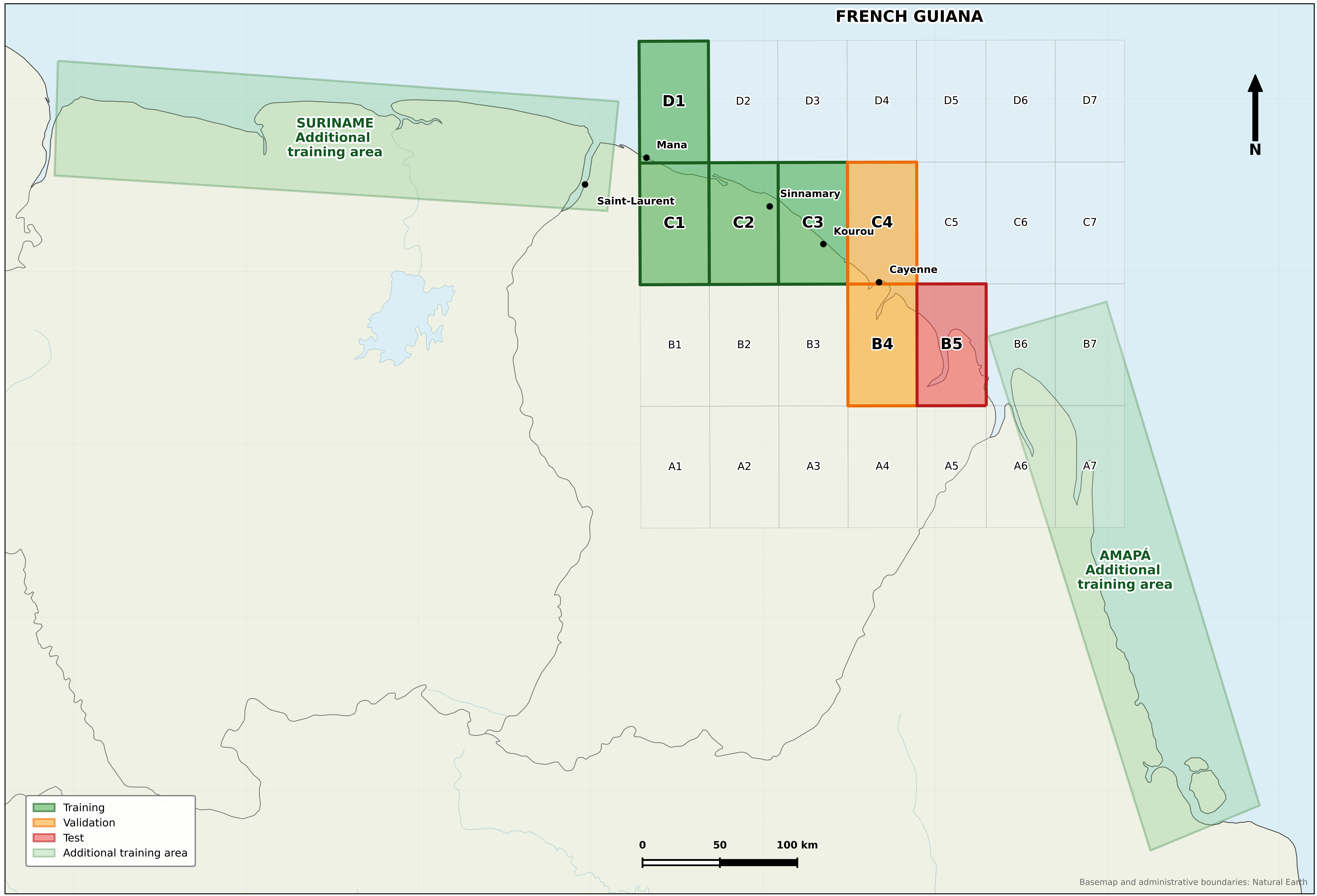}
    \caption{
    Spatial partitionning of the study area.
    }
    \label{fig:spatial_split}
\end{figure}

\paragraph{Study area}
The study focuses on the Atlantic coast of French Guiana, where mangrove ecosystems undergo rapid spatial changes driven by alternating phases of sediment accretion and coastal erosion. French Guiana constitutes the main study area, while additional observations from neighboring Suriname and Amapá State (Brazil) are used during training. (Figure~\ref{fig:spatial_split}).

\paragraph{Data preprocessing} We use Sentinel-1 (S1) Ground Range Detected (GRD) radar imagery and Sentinel-2 (S2) Level-2A multispectral imagery acquired during 2023 from the Copernicus Data Space Ecosystem and Microsoft Planetary Computer. S1 provides VV and VH polarizations while 12 S2 spectral bands (B01-B09, B8A, B11, and B12) are retained. Reference mangrove maps derived from 50~cm Pléiades imagery and available as vector and raster data provide the supervision labels. In total, 318 S1 and 781 S2 products were collected over French Guiana, with an additional 650 S1 and 2,800 S2 products covering Suriname and Amapá to enrich the training dataset.

S1 products were radiometrically calibrated, geometrically corrected using Range Doppler Terrain Correction, reprojected to UTM 22N (EPSG:32622) and converted to decibels. S2 bands were resampled to
10~m and spatially aligned with the Sentinel-1 grid. Cloud-contaminated pixels were identified using the S2 Scene Classification Layer (SCL) and OmniCloudMask\cite{wright2025}.

Because of persistent cloud cover, cloud-free observations were aggregated into quarterly temporal composites. For each pixel, the median of the available valid observations was computed within four three-month periods.

\paragraph{Dataset construction}
The harmonized S1 and S2 composites were divided into non-overlapping $128 \times 128$ pixel patches and assigned to a $7 \times 4$ spatial grid to limit spatial leakage (Fig.~\ref{fig:spatial_split}). Patches from cells C1, C2, C3, and D1 were used for training, C4 and B4 for validation, and B5 for testing. The remaining cells were excluded since they do not contain mangrove.

\paragraph{Models and Training}
We evaluated four segmentation configurations to compare different pretrained encoder and segmentation architectures: (1) CROMA \cite{fuller2023} with a simple convolutional decoder, (2) CROMA \cite{fuller2023} with UPerNet \cite{xiao2018}, (3) ResNet50 encoder pretrained on BigEarthNet \cite{clasen2025refinedbigearthnet,hackel2024configilm} with DeepLabV3+ \cite{Chen2018}, and (4) the same ResNet50 encoder combined
with UPerNet.

These configurations were selected to evaluate the influence of both the pretrained representation and the segmentation decoder. They allow us to assess whether multimodal CROMA \cite{fuller2023} representations provide an advantage for mangrove segmentation compared with a ResNet50 encoder pretrained on BigEarthNet \cite{clasen2025refinedbigearthnet,hackel2024configilm}, and whether the choice of decoder affects segmentation performance. Using UPerNet \cite{xiao2018} with both encoders provides a direct comparison of the pretrained representations under the same segmentation architecture.

In our experiments, the pretrained CROMA \cite{fuller2023} backbone was used to extract joint radar - optical representations, which were then provided either to a convolutional decoder or to UPerNet \cite{xiao2018}. UPerNet exploits multi-scale feature representations through a feature pyramid and a pyramid pooling module to recover both contextual and spatial information.

For comparison, the ResNet50 encoder pretrained on BigEarthNet \cite{clasen2025refinedbigearthnet,hackel2024configilm} received concatenated S1 and S2 inputs after adaptation to the expected input dimensions. The resulting multi-scale features were provided either to DeepLabV3+, which uses atrous convolutions for multi-scale contextual aggregation \cite{Chen2018}, or to UPerNet \cite{xiao2018}.

A visual representation of the four architectures is provided in Appendix~\ref{appendix:architectures}.

\section{Experiments and results}
\paragraph{Experimental setup}
The dataset comprised 37,630 training patches, 2,724 validation patches, and 2,256 test patches. The corresponding class distributions are reported in Table~\ref{tab:dataset_split}.

\begin{table}[!h]
    \centering
    \caption{Distribution of samples and mangrove pixels across dataset splits.}
    \label{tab:dataset_split}
    \begin{tabular}{lccc}
        \toprule
        \textbf{Set} & \textbf{Patches} &
        \textbf{Mangrove pixels} & \textbf{Non-mangrove pixels} \\
        \midrule
        Training   & 37,630 & 34.55\,\% & 65.45\,\% \\
        Validation & 2,724  & 37.66\,\% & 62.44\,\% \\
        Test       & 2,256  & 35.75\,\% & 64.25\,\% \\
        \bottomrule
    \end{tabular}
\end{table}

All models were trained on $128 \times 128$ pixel patches using AdamW with a batch size of 16 and a maximum of 100 epochs. Early stopping with a patience of 20 epochs was applied based on validation performance. Decision thresholds of 0.50, 0.55, 0.60, and 0.65 were evaluated for each model and the threshold with the highest F1-score and IoU was retained for reporting. Data augmentation consisting of horizontal and vertical flips and rotations of $90^\circ$, $180^\circ$, and $270^\circ$ was applied only to the training samples.

To address class imbalance, we combined focal binary cross-entropy and Dice losses \cite{yeung2022unified} with class weighting derived from the mangrove/non-mangrove distribution in the training set. Performance was assessed using precision, recall, F1-score, and intersection over union (IoU).

\paragraph{Mangrove mapping results}
Table~\ref{tab:final_models} summarizes the performance of the four segmentation configurations on the test set. 

\begin{table}[!h]
    \centering
    \caption{Test performance of the four segmentation configurations.}
    \label{tab:final_models}
    \small
    \setlength{\tabcolsep}{3pt}

    \begin{tabular}{lccccc}
        \toprule
        \textbf{Model} &
        \textbf{Thr.} &
        \textbf{Prec.} &
        \textbf{Recall} &
        \textbf{F1} &
        \textbf{IoU} \\
        \midrule

        CROMA + CNN
        & 0.65 & 0.824 & 0.864 & 0.844 & 0.730 \\

        CROMA + UPerNet
        & 0.50 & \textbf{0.906} & \textbf{0.938} &
        \textbf{0.922} & \textbf{0.855} \\

        ResNet50 + DeepLabV3+
        & 0.60 & 0.873 & 0.930 & 0.901 & 0.820 \\

        ResNet50 + UPerNet
        & 0.65 & 0.885 & 0.937 & 0.910 & 0.835 \\

        \bottomrule
    \end{tabular}
\end{table}

CROMA + UPerNet achieved the best overall performance, with an F1-score of 0.922 and an IoU of 0.855. Compared with CROMA + CNN, it achieved a 9.2\% relative improvement in F1-score and a 17.1\% relative improvement in IoU, showing that the segmentation decoder plays an important role even when the same pretrained representation is used. With UPerNet fixed as the decoder, CROMA also achieved a higher IoU than the ResNet50 representation pretrained on BigEarthNet (0.855 versus 0.835).

\begin{figure}[!h]
    \centering
    \includegraphics[width=\linewidth]{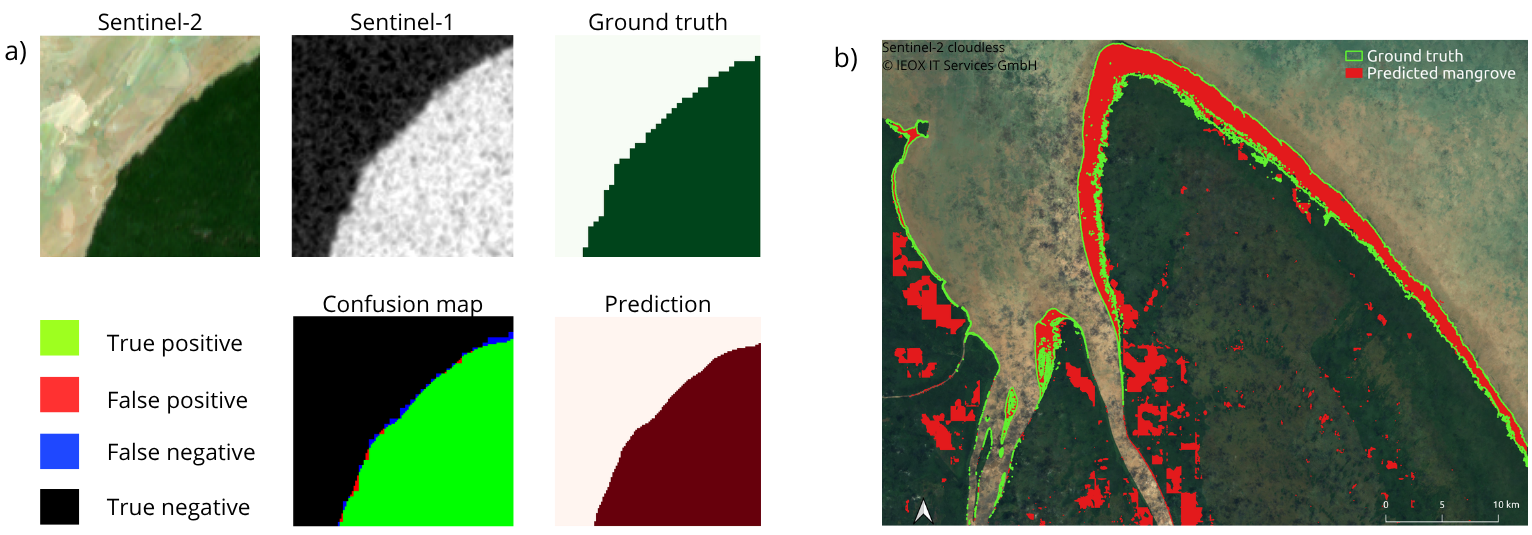}
    \caption{Results obtained with CROMA + UPerNet. 
    a) Patch-level segmentation showing Sentinel-1/2 quarterly composites, 
    reference mask, prediction, and pixel-wise confusion map. 
    b) Test-area prediction, with predicted mangrove pixels in red and 
    Pléiades-derived reference contours in green.}
    \label{fig:qualitative_map}
\end{figure}

Figure~\ref{fig:qualitative_map} provides a qualitative assessment of the model predictions at both patch and test-area scales. At the patch scale, the predicted mangrove extent closely matches the reference, with errors mainly concentrated along the boundary. The reconstruction over the complete test area provides a broader assessment of the spatial behavior of the model.

At this larger scale, most coastal mangrove formations are consistently identified. False positives are mainly associated with inland forested areas, suggesting that spectral and radar similarities between mangrove and terra firme forests remain a source of confusion for the model.

\section{Discussion and conclusion}
The experiments highlight the importance of data construction and training strategy for mangrove segmentation. Early experiments based on annual S1 and S2 composites and a strongly imbalanced dataset resulted in poor segmentation performance with a best IoU below 0.17. The experimental protocol was progressively refined by using quarterly composites, reducing class imbalance, introducing data augmentation and a Focal-Dice loss with class weighting and enriching the training set with data from Suriname and Amapá. Together, these changes led to the substantially higher performance reported here, with a best IoU of 0.855. The experimental progression is summarized in Appendix~\ref{appendix:experimental_evolution}.

Architecture comparisons show that segmentation performance depends on both the pretrained representation and the decoder with CROMA + UPerNet providing the best combination in our experiments. Spatial evaluation nevertheless reveals false positives over inland forests, indicating remaining confusion between mangroves and terrestrial forests. To address this limitation, a further experiment will incorporate a terrestrial forest mask during training to help the model better distinguish these two vegetation types.

These results provide a first step toward transferring information from very-high-resolution reference products to Sentinel observations. Future work will extend the framework beyond binary mangrove delineation toward habitat characterization and ultimately, above-ground biomass and carbon-stock estimation. Multi-year and cross-region evaluation will also be needed to assess the robustness and generalization of the approach. These developments could contribute to scalable monitoring of mangrove extent, condition and carbon-related properties for blue-carbon assessment, conservation and climate mitigation and adaptation.

\begin{ack}
This work was conducted at UMR ESPACE-DEV, IRD, Montpellier, France, as part of the ESA Coastal Blue Carbon and CNES ANIMALS (Artificial iNtellIgence for the Mapping of MAngrove at Large Scale) projects. 

This work was performed using HPC resources from GENCI-IDRIS (Grant 2025-AD011016948). This work was funded by University of Montpellier DEMON (2025-2027) project.
\end{ack}

\bibliographystyle{unsrt}
\bibliography{biblio}

\newpage
\section{Appendix}
\begin{figure}[htbp] \centering \resizebox{\textwidth}{!}{%
\begin{tikzpicture}[ font=\small, >=Latex, mainbox/.style={ rectangle, rounded corners=5pt, draw=black!45, fill=white, minimum width=4.15cm, minimum height=12.8cm, inner sep=0pt }, titlebox/.style={ rectangle, rounded corners=4pt, draw=black!35, fill=white, minimum width=3.75cm, minimum height=1.25cm, align=center, text width=3.55cm, font=\small\bfseries }, problem/.style={ rectangle, rounded corners=4pt, draw=orange!65!black, fill=orange!8, minimum width=3.65cm, minimum height=2.05cm, align=left, text width=3.25cm, inner sep=6pt, font=\scriptsize }, method/.style={ rectangle, rounded corners=4pt, draw=blue!45!black, fill=blue!5, minimum width=3.65cm, minimum height=2.35cm, align=left, text width=3.25cm, inner sep=6pt, font=\scriptsize }, result/.style={ rectangle, rounded corners=4pt, draw=green!50!black, fill=green!7, minimum width=3.65cm, minimum height=2.05cm, align=center, text width=3.25cm, inner sep=6pt, font=\scriptsize }, arrow/.style={ -{Latex[length=5mm,width=3mm]}, line width=1.8pt, draw=black!60 }, progression/.style={ -{Latex[length=4mm,width=2.5mm]}, line width=1.5pt, draw=green!50!black } ] \coordinate (C1) at (0,0); \coordinate (C2) at (5.1,0); \coordinate (C3) at (10.2,0); \coordinate (C4) at (15.3,0); \coordinate (C5) at (20.4,0); 

 \draw[progression] (-1.5,6.75) -- (22.1,6.75); \node[ fill=white, inner sep=4pt, font=\small\bfseries, text=green!50!black ] at (10.2,6.75) {Progressive refinement of the experimental strategy}; \node[mainbox] (box1) at (C1) {}; \node[titlebox] at ($(C1)+(0,5.25)$) { \textbf{Sentinel-1 only}\\[2mm] SAR data };  \node[method] at ($(C1)+(0,1.55)$) { \textbf{Method}\\[1mm] $\bullet$ CROMA + CNN\\ $\bullet$ Dice Loss + BCE\\ $\bullet$ Decision threshold tests\\ $\bullet$ Batch-size tests}; \node[result] at ($(C1)+(0,-2.0)$) { \textbf{\textcolor{green!45!black}{Best result}}\\[2mm] F1-score = \textbf{0,5886}\\ IoU = \textbf{0,4171} }; 

\node[mainbox] (box2) at (C2) {}; \node[titlebox] at ($(C2)+(0,5.25)$) { \textbf{Sentinel-2 only}\\[2mm] Optical data }; \node[problem] at ($(C2)+(0,3.05)$) { \textbf{\textcolor{orange!70!black}{Identified limitations}}\\[1mm] $\bullet$ Cloud contamination\\ $\bullet$ Missing pixels\\ $\bullet$ Many areas unsuitable\\ \hspace*{2mm}for training }; \node[method] at ($(C2)+(0,-0.15)$) { \textbf{Method}\\[1mm] $\bullet$ CROMA + CNN\\ [1mm] $\bullet$ Dice Loss + BCE\\ [1mm] $\bullet$ Two thresholds tested }; \node[result] at ($(C2)+(0,-3.45)$) { \textbf{\textcolor{green!45!black}{Best result}}\\[2mm] F1-score = \textbf{0,4089}\\ IoU = \textbf{0,2570} }; \draw[arrow] ($(C1)+(2.15,0)$) -- ($(C2)+(-2.15,0)$); 

\node[mainbox] (box3) at (C3) {}; \node[titlebox] at ($(C3)+(0,5.25)$) { \textbf{Annual composites}\\[2mm] Sentinel-1 + Sentinel-2 }; \node[problem] at ($(C3)+(0,2.50)$) { \textbf{\textcolor{orange!70!black}{Identified limitations}}\\[1mm] $\bullet$ Limited training data\\[1mm] $\bullet$ High proportion of non-mangrove pixels\\ [1mm] $\bullet$ Strong class imbalance }; \node[method] at ($(C3)+(0,-0.90)$) { \textbf{Adaptations}\\[1mm] $\bullet$ Focal Dice Loss\\ [1mm] $\bullet$ Four architectures\\ [1mm] $\bullet$ Thresholds 0.50 -- 0.65 }; \node[result] at ($(C3)+(0,-4.40)$) { \textbf{\textcolor{green!45!black}{Best result}}\\[2mm] ResNet50 + UPerNet\\ F1-score = \textbf{0,2901}\\ IoU = \textbf{0,1697} }; \draw[arrow] ($(C2)+(2.15,0)$) -- ($(C3)+(-2.15,0)$);

\node[mainbox] (box4) at (C4) {}; \node[titlebox] at ($(C4)+(0,5.25)$) { \textbf{Quarterly composites}\\[2mm] Sentinel-1 + Sentinel-2 }; \node[problem] at ($(C4)+(0,2.50)$) { \textbf{\textcolor{orange!70!black}{Objectives}}\\[1mm] $\bullet$ Increase training data\\[1mm] $\bullet$ Better preserve temporal\\ [1mm] $\bullet$ Reduce class imbalance\\ [1mm] $\bullet$ Improve segmentation }; \node[method] at ($(C4)+(0,-0.85)$) { \textbf{Adaptations}\\[1mm] $\bullet$ Quarterly composites\\ [1mm] $\bullet$ Fewer negative patches\\ [1mm] $\bullet$ Data augmentation\\ [1mm] $\bullet$ Weighted Focal-Dice loss }; \node[result] at ($(C4)+(0,-4.45)$) { \textbf{\textcolor{green!45!black}{Best result}}\\[2mm] DeepLabV3+\\ F1-score = \textbf{0,6294}\\ IoU = \textbf{0,4593} }; \draw[arrow] ($(C3)+(2.15,0)$) -- ($(C4)+(-2.15,0)$);

\node[ mainbox, draw=green!50!black ] (box5) at (C5) {}; \node[titlebox] at ($(C5)+(0,5.25)$) { \textbf{Data Enrichment}\\[2mm] French Guiana + Suriname + Amapá }; \node[problem] at ($(C5)+(0,2.30)$) { \textbf{\textcolor{green!45!black}{Dataset enrichment}}\\[1mm] $\bullet$ Increased geographic\\diversity\\ [1mm] $\bullet$ Further reduction of negative patches\\ [1mm] $\bullet$ Improved class balance }; \node[method] at ($(C5)+(0,-1.60)$) { \textbf{Final configuration}\\[1mm] $\bullet$ Quarterly composites\\ [1mm] $\bullet$ Data augmentation\\ [1mm] $\bullet$ Weighted Focal-Dice loss\\ [1mm] $\bullet$ Further reduction of negative patches}; \node[ result, draw=green!65!black, fill=green!12 ] at ($(C5)+(0,-5.00)$) { \textbf{\textcolor{green!45!black}{Best result}}\\[2mm] \textbf{CROMA + UPerNet}\\ F1-score = \textbf{0,9218}\\ IoU = \textbf{0,8549} }; \draw[arrow] ($(C4)+(2.15,0)$) -- ($(C5)+(-2.15,0)$);

\end{tikzpicture}%
} \caption{Evolution of the experimental strategy from single-modality
Sentinel-1 and Sentinel-2 experiments to the final geographically enriched
multimodal training configuration.} \label{fig:evolution_experimentations} \end{figure}
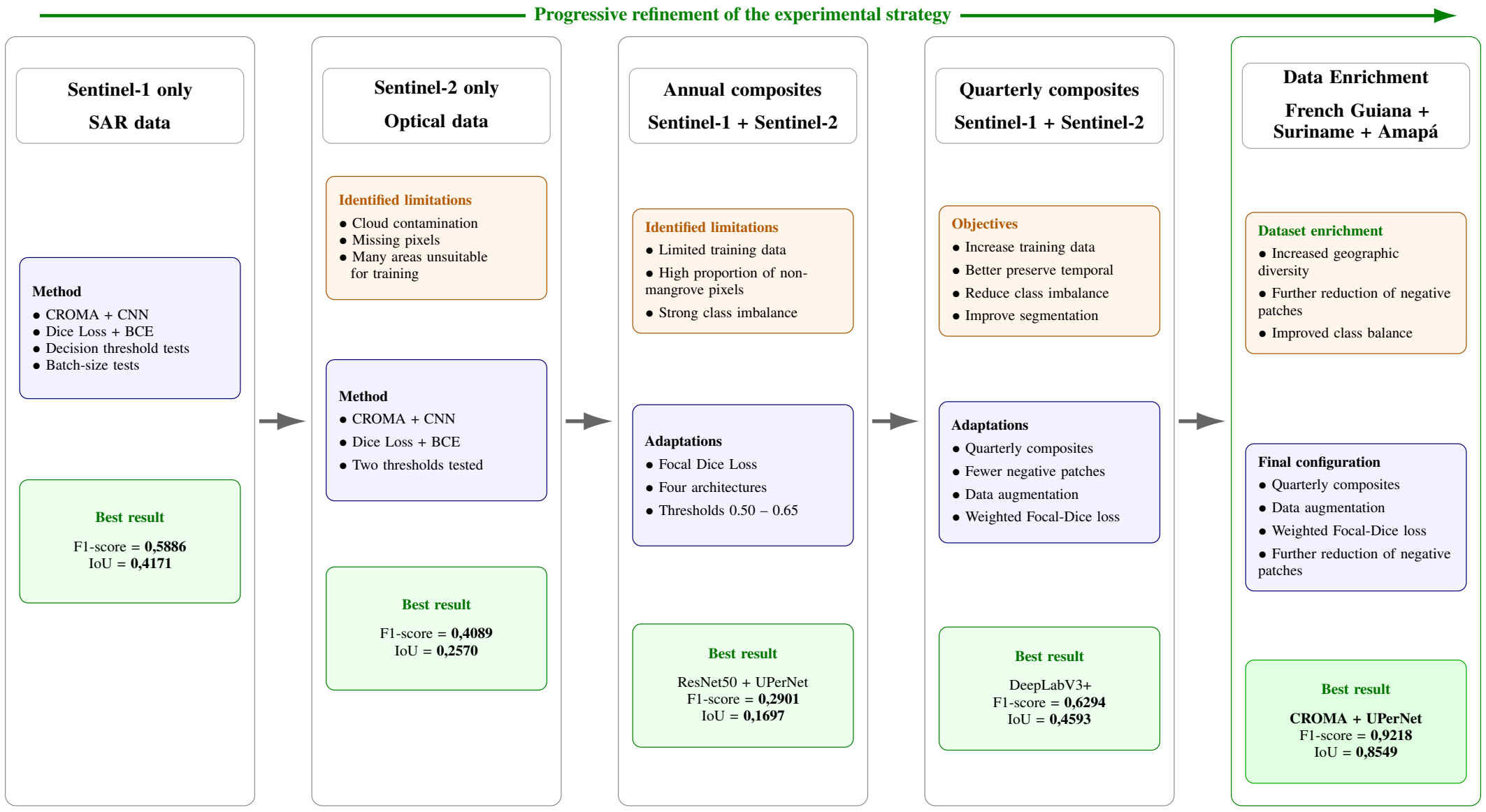

\paragraph{Evolution of the experimental strategy}
\label{appendix:experimental_evolution}
Figure~\ref{fig:evolution_experimentations} summarizes the successive
experimental configurations investigated during the study, from the initial
single-modality experiments to the final geographically enriched training
dataset. Each stage motivated subsequent changes to the data representation, class-balancing strategy, or training protocol.

\paragraph{Segmentation architectures}
\label{appendix:architectures}
The four segmentation configurations evaluated in this study are illustrated below. The diagrams show the processing of Sentinel-1 and Sentinel-2 inputs, the pretrained encoder or backbone, the segmentation decoder, and the resulting mangrove segmentation map.

\begin{figure*}[!h]
    \centering
    \includegraphics[width=\textwidth]{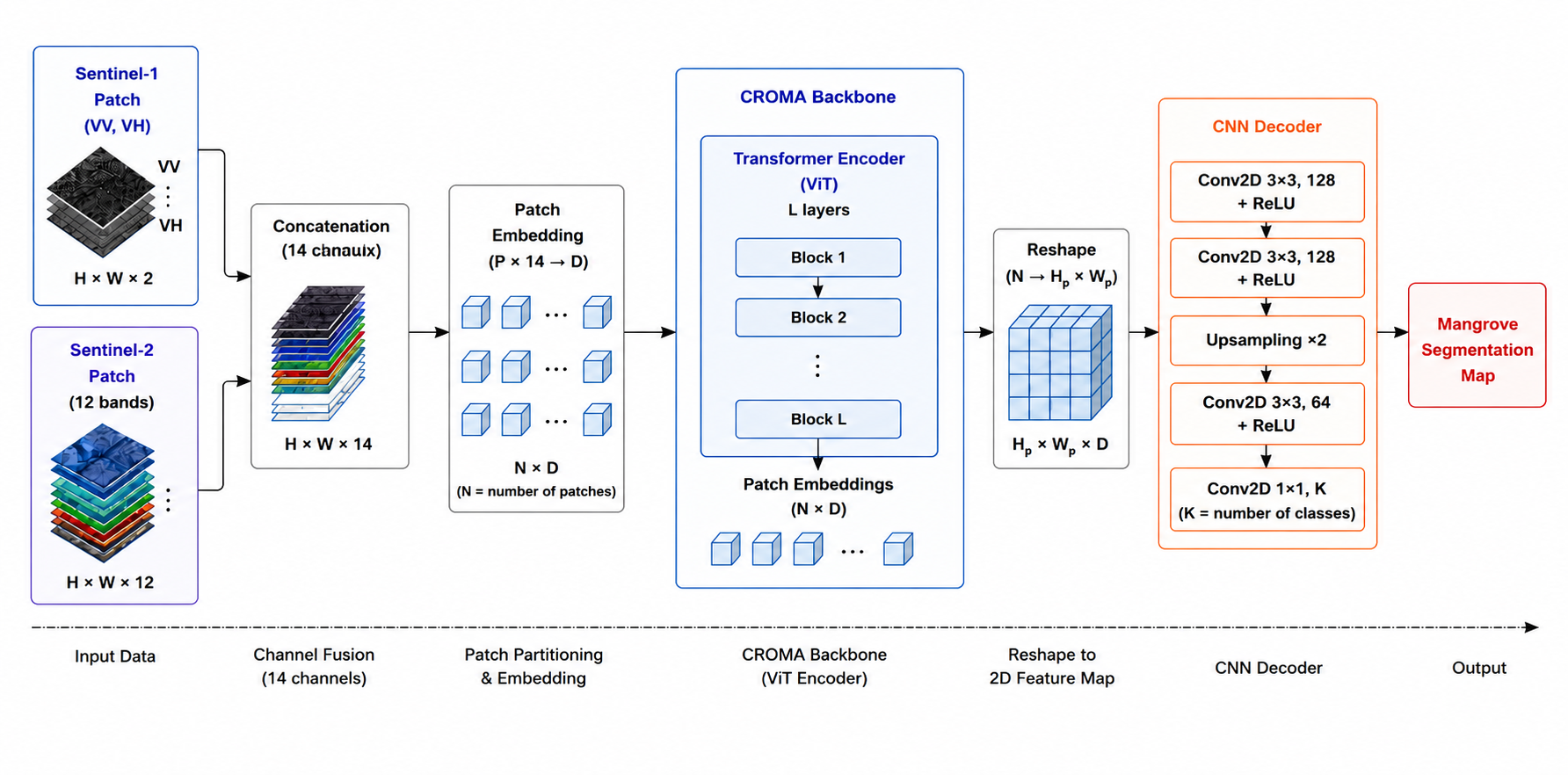}
    \caption{Architecture of the CROMA + CNN configuration. Joint S1 and S2 representations extracted by the pretrained CROMA backbone
    are reshaped into a 2D feature map and processed by a convolutional
    decoder to produce the mangrove segmentation map.}
    \label{fig:arch_croma_cnn}
\end{figure*}

\begin{figure*}[!h]
    \centering
    \includegraphics[width=\textwidth]{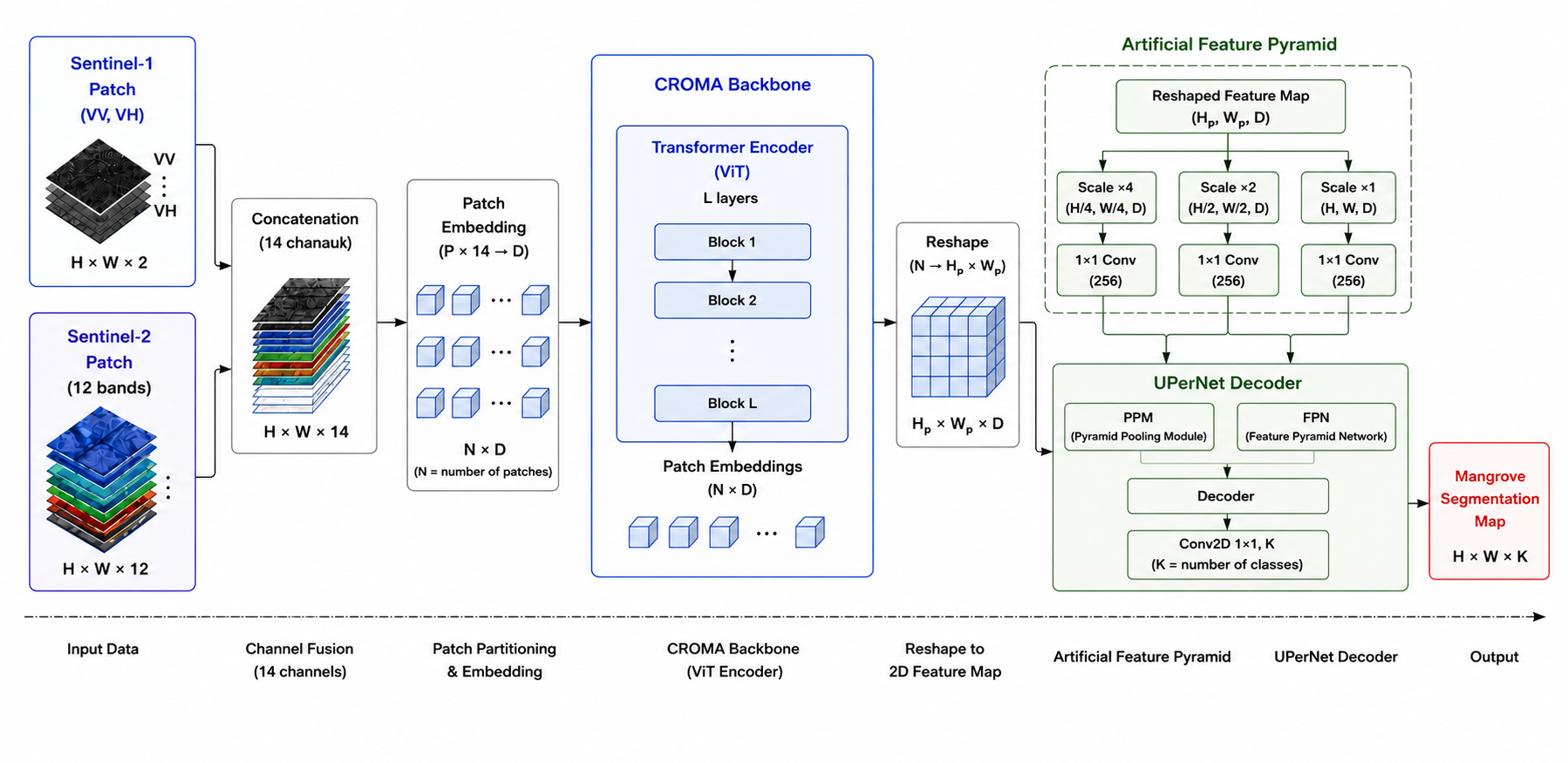}
    \caption{Architecture of the CROMA + UPerNet configuration. Features
    extracted from S1 and S2 by the pretrained CROMA backbone are provided to the UPerNet decoder for multi-scale mangrove segmentation.}
    \label{fig:arch_croma_upernet}
\end{figure*}

\begin{figure*}[!h]
    \centering
    \includegraphics[width=\textwidth]{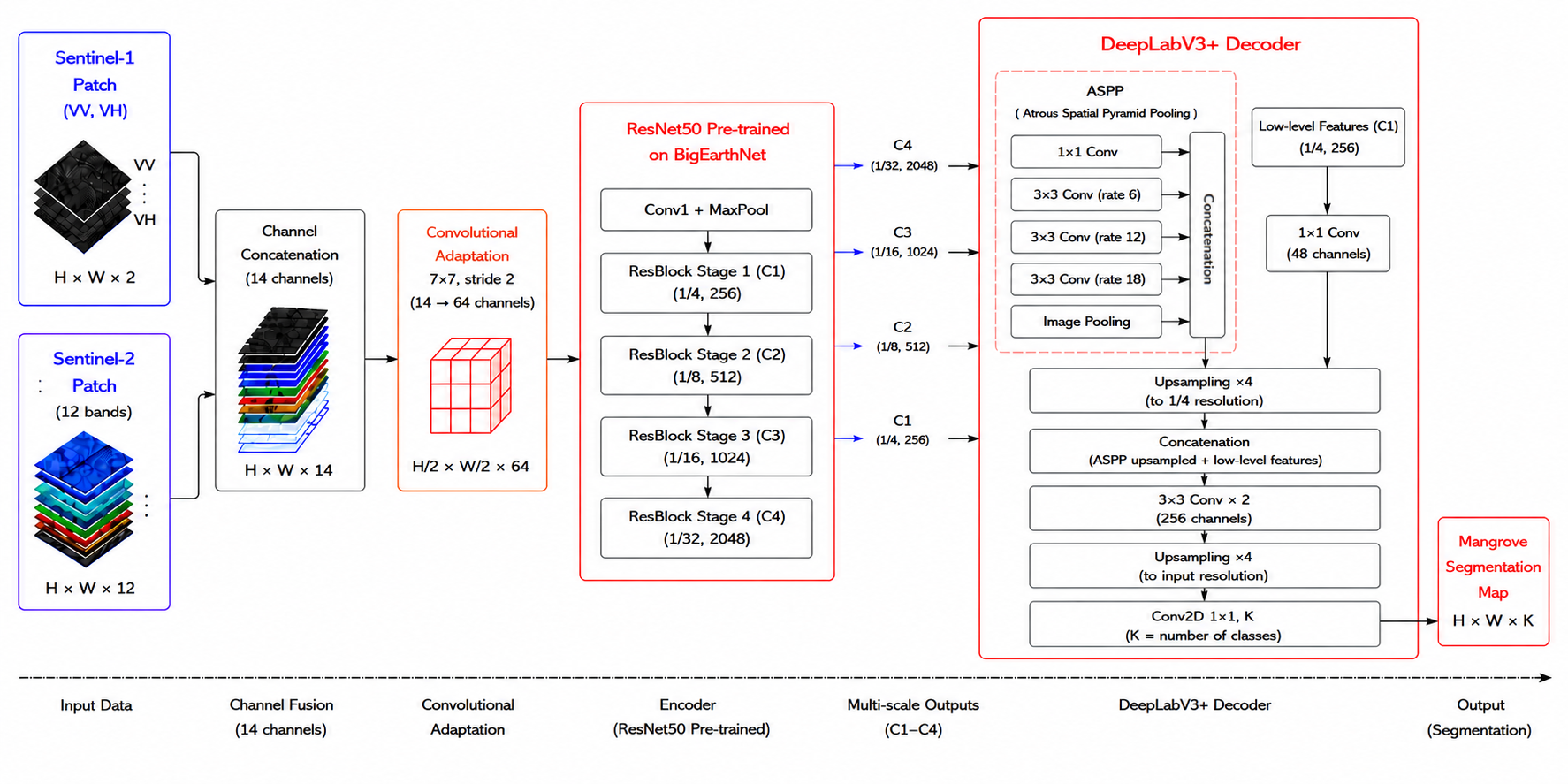}
    \caption{Architecture of the ResNet50 + DeepLabV3+ configuration. S1 and S2 inputs are combined and processed by the ResNet50 encoder pretrained on BigEarthNet while DeepLabV3+ performs multi-scale feature aggregation and produces the segmentation map.}
    \label{fig:arch_resnet_deeplab}
\end{figure*}

\begin{figure*}[!h]
    \centering
    \includegraphics[width=\textwidth]{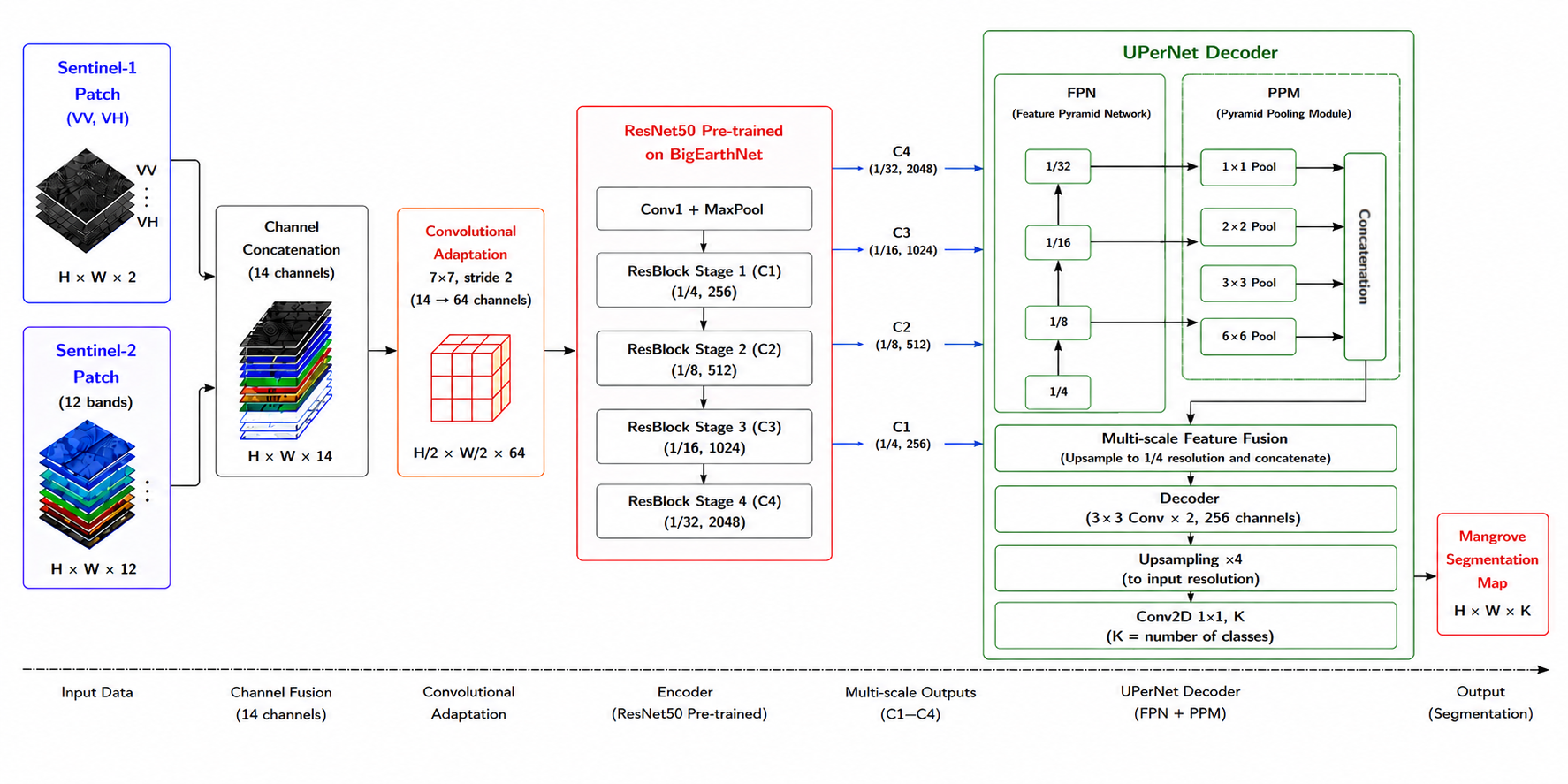}
    \caption{Architecture of the ResNet50 + UPerNet configuration. S1 and S2 inputs are combined and processed by the ResNet50 encoder pretrained on BigEarthNetand and the resulting multi-scale features are
    provided to UPerNet for mangrove segmentation.}
    \label{fig:arch_resnet_upernet}
\end{figure*}

\end{document}